# Applying HCAI in Developing Effective Human-AI Teaming: A Perspective from Human-AI Joint Cognitive Systems

Wei Xu and Zaifeng Gao

Zhejiang University, China

**Insights**
- AI has led the emergence of a new form of human-machine relationship: human-AI teaming (HAT), a paradigmatic shift in human-AI systems.
- We must follow a human-centered AI (HCAI) approach when applying HAT as a new design paradigm.
- We propose a conceptual framework of human-AI joint cognitive systems (HAIJCS) to represent and implement HAT for developing effective human-AI teaming.

In this article, we extend our previous work of exploring levels of collaboration between artificial intelligence and humans, enabling the development of human-centered AI (HCAI) systems [1,2]. Researchers have used human-AI teaming (HAT) as a new paradigm to develop AI systems [3]. HAT recognizes that AI will function as a teammate instead of simply a tool in collaboration with humans. Effective human-AI teams need to be capable of taking advantage of the unique abilities of both humans and AI while overcoming the known challenges and limitations of each member, augmenting human capabilities, and raising joint performance beyond that of either entity. The *National Artificial Intelligence Research and Development Strategic Plan 2023 Update* has recognized that research programs focusing primarily on the independent performance of AI systems generally fail to consider the functionality that AI must provide within the context of dynamic, adaptive, and collaborative teams and calls for further research on human-AI teaming and collaboration [4]. There has been debate, however, about whether AI can work as a teammate with humans [3]. The primary concern is that adopting the "teaming" paradigm contradicts the human-centered AI (HCAI) approach, resulting in humans losing control of AI systems [5].

This article further analyzes the HAT paradigm and its debates. Specifically, we elaborate on our proposed conceptual framework of human-AI joint cognitive systems (HAIJCS) and apply it to represent HAT under the HCAI umbrella [6]. We believe that HAIJCS may help adopt human-AI teaming while enabling human-centered AI. The implications and future work for HAIJCS are also discussed.

**Unique Characteristics of AI Technology**
There are significant differences between non-AI computing systems—the focus of conventional HCI work—and AI systems from an HCI perspective [1,2]. Herein, we define AI systems as AI-based computing systems that replicate humanlike cognitive capabilities such as sensing, learning, and reasoning, allowing the systems to operate autonomously in specific environments. When interacting with humans, non-AI systems primarily work as assistive tools to support humans driven by the conventional "stimulus response" interactions; the machines rely on fixed logical rules and algorithms to respond to human instructions. Today, the relationship between humans and AI systems goes beyond mere "interaction"; it is driven by the unique features, such as two-way, active, shared, complementary, goal-driven, and predictable characteristics in AI systems [2].

Therefore, the development of AI systems can take advantage of its unique capabilities while overcoming the known challenges and limitations of humans and AI systems. From the perspective of collaboration, there is limited human-machine collaboration for non-AI systems compared to potentially more-effective human-AI collaboration. The unique capabilities of AI build the foundation for human-AI teaming.

**A Paradigmatic Shift: From Interaction to Teaming**

Historically, the evolution of human-machine relationships has been driven by technology. As shown in Figure 1, the human-machine relationship has evolved from the "humans adapt to machines" paradigm that preceded World War II to the "machines adapt to humans" paradigm after the war. In the computer era, it evolved to human-computer interaction, where non-AI computing systems primarily work as assistive tools supporting humans. As we enter the AI era, the human-machine relationship can be characterized by HCI + HAT. AI systems work as assistive tools, but they could also work as collaborative teammates with humans, thus playing dual roles.

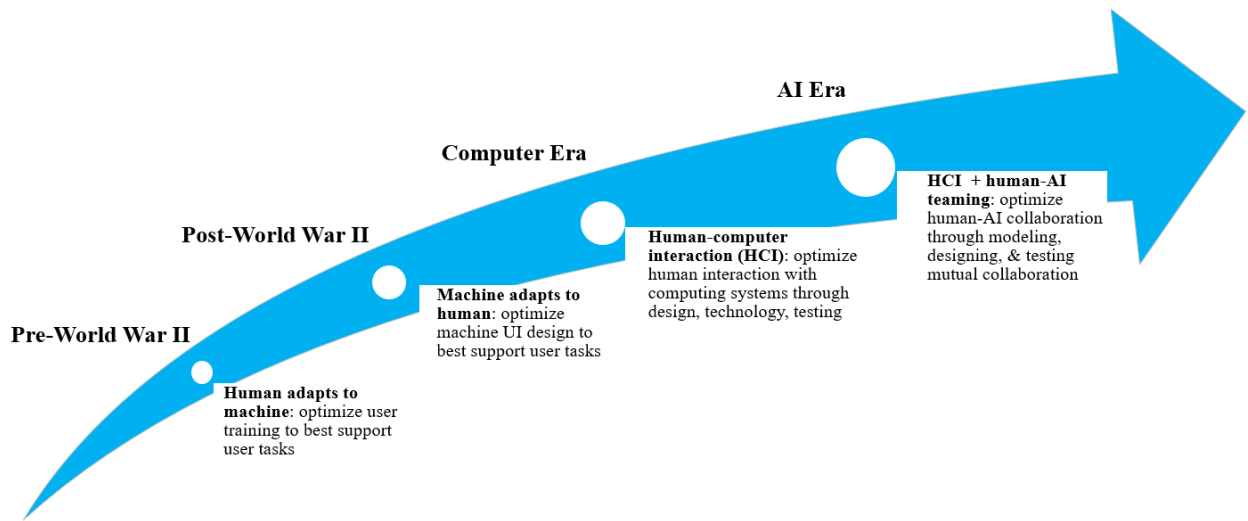

Figure 1. The paradigmatic shifts in the human-machine relationship across eras.

Academic communities have not yet reached a consensus on whether AI can become a teammate in collaboration with humans. Some researchers argue against using the "teammate" design metaphor for AI; they are concerned that adopting HAT may result in humans losing control of AI systems, thereby contradicting HCAI. Another argument is that current AI technology cannot provide the capabilities of real teammates. Instead, it is suggested that we use design metaphors such as "super tools," as these could more effectively guide developers to achieve the design goal of HCAI [5].

Considering AI as a teammate to human operators goes back several decades. Recent NASA work has posited major tenets for HAT, and the National Academies outlined comprehensive research needs for HAT in its 2021 report [3]. The recent *National Artificial Intelligence Research and Development Strategic Plan Update* called for future research on HAT-based collaboration [4].

HAT is a rapidly progressing research area, especially in the human factors community. Some believe that there is no contradiction between HAT and HCAI. For instance, the contributors to the National Academies report strongly believe that adopting HAT has essential benefits. By treating humans and AI as teammates, the value of team interaction in producing superior performance over independent individuals can be explored. Also, the teaming paradigm does not necessarily imply that these agents are equivalent in agency, capability, responsibility, or authority, similar to human-human teams. The literature on human-robot collaboration generally accepts that humans should be in charge of teams for ethical and practical reasons [3].

From a long-term perspective, while current AI systems are far from meeting the criteria for effective teammates, there is value in emphasizing the criteria as teammates and working toward building AI systems that meet them. In fact, the military already has a long history of humans collaborating with nonhumans [3]. People who support HAT agree with the concerns raised and support the need for HCAI when developing it. HAT must be human-centric, and an AI teammate should not take over control of the system from a human [2]. Also, more than three decades of research literature on teamwork provide extensive guidance for effective teaming and show that designing an AI system capable of working with teammates can increase human-centeredness [3]. Ultimately, an effective human-AI team augments human capabilities and raises performance beyond that of either entity.

As history has proved, some people initially hesitated to adopt the emerging metaphor of "human-computer interaction" in the early stage of the computer era. We have since witnessed another emerging form of human-machine relationship ushered in by AI technology, resulting in a paradigmatic shift for new design thinking in the development of AI systems. While the debates about the teammate metaphor continue, humans and AI are collaborating more as technology advances—the recent development of ChatGPT is an excellent example of this. How then can we embrace this new form of human-machine relationship and address the valid concerns raised about HAT from the HCAI perspective?

**A Conceptual Framework of Human-AI Joint Cognitive Systems for Human-AI Teaming**
HCI professionals used to model the interactions between humans and computing systems, which were primarily aimed at human interaction with non-AI systems. Based on Erik Hollnagel and David Woods's joint cognitive systems theory, Mica Endsley's situation awareness cognitive engineering theory, and the agent theory widely used in AI/CS communities, we propose a conceptual framework of joint cognitive systems to represent HAT (Figure 2) [6].

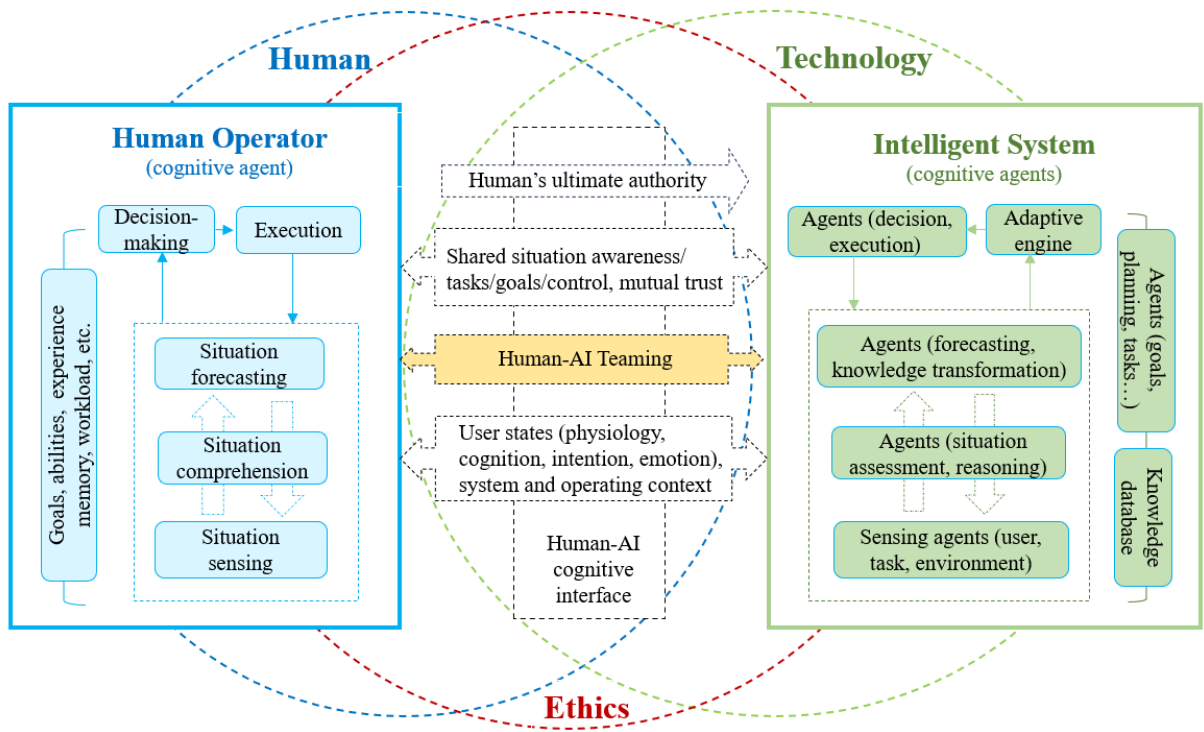

Figure 2. The conceptual framework of human-AI joint cognitive systems (HAIJCS).

As shown in Figure 2, HAIJCS regards AI systems (with one or more AI agents) as cognitive agents that can perform specific tasks based on their autonomous capabilities. Therefore, a human-AI team can be characterized as a joint cognitive system in which the two cognitive agents collaborate. Human biological intelligence and machine intelligence achieve intelligent complementarity through their collaboration.

HAIJCS applies Endsley's situation awareness theory to characterize the information-processing mechanism of the two cognitive agents. The model represents how human users perceive and understand the current state of the environment and predict future situational states through Endsley's model of three levels of information processing, as illustrated by the information processing mechanism in both data-driven and goal-driven approaches. It also includes cognitive interactions between situation awareness and memory, experience, knowledge, and so on. HAIJCS mimics the architecture used by the human agent to represent the information processing mechanism for the cognitive agents on the AI side. Both cognitive agents collaborate through a human-AI cognitive interface driven by multimodal technology while maintaining shared situation awareness, task, goal, trust, decision making, and control under the umbrella of HCAI (e.g., human's ultimate authority).

HAIJCS represents a new perspective for further understanding and implementing HAT, eventually developing effective human-AI teaming. First, HAIJCS applies human-centered AI to facilitate the implementation of human-AI teaming. The HCAI approach, as originally proposed by us in 2019, seeks synergy among the three aspects: human, technology, and ethics (see the three circles with dotted lines in Figure 2). If the development of human-AI systems only considers humans and technology while ignoring ethics (e.g., fairness, privacy), such a human-AI team will eventually harm humans. If it only considers humans and ethics without pursuing advanced AI technology, the system lacks technical feasibility for implementation and sustainability in technology. On the other hand, if it only considers technology and ethics but ignores humans (e.g., user needs and experience), the system may not be useful and usable. Thus, the overlapping area obtained after thoroughly considering all three factors is the space for the success of human-centered human-AI systems. As illustrated, HAIJCS takes HCAI as an umbrella concept to promote a design philosophy. Humans are the leaders in the human-AI teams with ultimate authority; developing human-AI systems is to augment human capabilities.

Second, HAIJCS promotes AI as a cognitive agent to be a qualified teammate for human-AI collaboration, unlike the conventional approach that regards machines as tools to assist humans. An agent is considered a programmable object with autonomous characteristics in the CS/AI community; HAIJCS provides new design thinking to "promote" AI agents in the collaboration context from a long-term perspective. Although current AI systems are far from meeting the criteria for effective teammates, considering them as cognitive agents may further promote agency technology toward building more powerful human-centered AI systems. HAIJCS mimics the method with the human agent to conceptually characterize the information processing mechanism for AI agents, which may meaningfully define a conceptual reference architecture for developing AI agents. The autonomous characteristics of AI agents, such as sensing, reasoning, decision making, and execution, may be implemented by leveraging the human information processing mechanism widely developed by cognitive science. Conventional approaches for building AI systems are to consider the capabilities of humans and AI separately. Now we need to think of a new way to develop AI agents that can work as "qualified cognitive agents" to collaborate with humans while developing human-centered HAT.

Third, HAIJCS suggests that humans and AI are not two independent parts but function as a whole, namely human-AI teaming. The performance of the entire cognitive system depends on the degree of synergy between the two parts, not just the performance of one. HAIJCS fosters the need to consider each team member's interrelated roles and emphasizes the value of *team interactions* for boosting their joint performance, instead of conventional HCI approaches that focus on designing the stimulus-response interaction.

Last, HAIJCS represents HAT as a collaborative relationship between the two cognitive agents, encouraging us to mimic mature human-human teamwork to develop effective human-AI teaming. By leveraging human-human teaming work, we may explore alternative ways to improve the human-AI joint performance by studying, modeling, designing, building, and testing the collaborative relationship between the two cognitive agents. With such a mindset change, AI agents are no longer "traditional" machine agents. For instance, as illustrated in Figure 2, we may adopt mature human-human teaming approaches to model shared situation awareness, trust, decision making, and control between the two agents through multidisciplinary collaboration across the communities of AI/CS, HCI, human factors, cognitive science, and more.

**Implications for Application**

We are currently exploring the application of HAIJCS in the advanced autonomous vehicle (AV) domain. AV can be considered a joint cognitive system where human drivers and AI-based in-vehicle agents collaboratively complete driving tasks. Although HCI professionals are involved in developing AVs, frequent accidents compelled us to explore new design approaches. AVs at or below Level 4, as defined by SAE J3014, still require human drivers to participate in driving activities. HAIJCS can help us study alternative design approaches that seek human-vehicle collaborative codriving solutions (see Table 1).

| Aspects of Analysis | Current Approaches | Implications of HAIJCS |
|---|---|---|
| Classification of automated driving | SAE J3014 defines six numerical classifications on levels of automated driving. This technology-centered taxonomy emphasizes what AVs can do regarding a binary function allocation between human driver and AV. | HAIJCS advocates the collaboration between driver and AV, informing the design of a more collaborative relationship and dynamic functional allocation. |
| Human-machine relationship | These siloed approaches focus on human and AI agents in silos. The AV is considered a tool to assist or replace human driving, while human drivers primarily play supervisory control. | Driver and AV are considered cognitive agents for collaboration. It may free the driver from the unsustainable supervisory role to a collaborative agency through a "human-in-the-loop" design and shared responsibilities/intentionality. |
| Interaction paradigm | Unidirectional interaction paradigms are primarily deployed based on the traditional stimulus-response unidirectional human-machine interaction. | HAIJCS provides a useful interaction paradigm promoting bidirectional interaction with shared responsibilities between driver and AV. It promotes in-vehicle collaborative cognitive interfaces. |
| Design philosophy | Human-centered design is typically not considered. The leadership role of driver-AV systems is unclear. | HAIJCS emphasizes the human-centered design approach and promotes humans as the leaders of the collaborative teams (at least for vehicles L4 and below) and the final controller of the system in an emergency (e.g., 5G remote control at L5). |
| Theoretical framework | There is a lack of theoretical frameworks and theories for effectively studying human-vehicle codriving. | HAIJCS encourages researchers to leverage and mimic human-human team theories and frameworks to design, model, and develop driver-AI teaming, such as mutual and shared situation awareness, trust, and control. |

Table 1. A comparative analysis of the current and HAIJCS approach in human-vehicle codriving.

Other domains for applying HAIJCS include smartphones that may work as collaborative teammates with consumers, intelligent robots that may work as collaborative teammates with users (e.g., patients, elderly people), and future intelligent civil flight decks where human captains and AI copilots will be collaborative teammates.

**Call for Multidisciplinary Collaboration**
The research on HAT is still immature, and more work is needed. Guided by HCAI, we must clarify the roles of humans in human-AI systems while further exploring HAT. HAIJCS may be one of the approaches to reach that goal. As a conceptual framework, HAIJCS also needs further exploration. We need to enhance the framework to better support and validate the HAT metaphor by applying it to various domains.

The successful development of a human-AI system that can function as a good teammate will require multidisciplinary collaboration across AI/CS, HCI, human factors, cognitive science, and other disciplines, such as an interdisciplinary development team, methods and tools, and improved metrics for measuring joint collaborative performance, as represented in the HAIJCS framework. Current work on HAT is focused on the human effectiveness component of human-AI teaming, specifically led by the human factors community [3]. AI/CS communities should become involved, for instance, in building the needed quantitative predictive models of human-AI performance.

There are values to adopting the HAT metaphor as long as HCAI is applied; the question is how we can effectively use HAT to develop human-centered AI systems. We believe that HAIJCS is an approach worthwhile to explore further. We may consider HAT an opportunity to implement HCAI rather than a barrier.

**Endnotes**

**Wei Xu** is a professor of human factors/HCI at Zhejiang University in China. He is an elected fellow of the International Ergonomics Association, the International Association for Psychological Science, and the Human Factors and Ergonomics Society. He received a Ph.D. in psychology with an emphasis on human factors/HCI and an M.S. in computer science from Miami University. His research interests include human-AI interaction, HCI, and aviation human factors. weixu6@yahoo.com

**Zaifeng Gao** is a professor of psychology/human factors in the Department of Psychology and Behavioral Sciences at Zhejiang University in China. He received a Ph.D. in psychology from Zhejiang University. His research interests include engineering psychology, autonomous driving, and cognitive psychology. zaifengg@zju.edu.cn